\documentclass[conference]{IEEEtran}
\IEEEoverridecommandlockouts

\usepackage{cite}
\usepackage{amsmath,amssymb,amsfonts}
\usepackage{algorithmic}
\usepackage{graphicx}
\usepackage{textcomp}
\usepackage{xcolor}
\usepackage{float}
\usepackage{booktabs}
\usepackage{amssymb}
\usepackage{color}
\usepackage{subfigure}
\usepackage{multirow}
\usepackage{array}
\usepackage{hyperref}
\usepackage{pdfpages}
\usepackage{tabularx}

\def\BibTeX{{\rm B\kern-.05em{\sc i\kern-.025em b}\kern-.08em
    T\kern-.1667em\lower.7ex\hbox{E}\kern-.125emX}}
\begin{document}

\title{Few-Shot Continual Learning for Activity Recognition in Classroom Surveillance Images
\thanks{ ${}^{\dag}$ Corresponding author (lfxu@uestc.edu.cn)} 
\thanks{ $^*$ Equal Contribution} 
}

\author{
    Yilei Qian$^*$, Kanglei Geng$^*$, Kailong Chen, Shaoxu Cheng, Linfeng Xu$^{\dag}$, Hongliang Li, Fanman Meng, Qingbo Wu \\
    \textit{School of Information and Communication Engineering} \\
    \textit{University of Electronic Science and Technology of China, Chengdu, China} \\
    \{yileiqian, kangleigeng, chenkailong, shaoxu.cheng\}@std.uestc.edu.cn, \{lfxu, hlli, fmmeng, qbwu\}@uestc.edu.cn
}

\maketitle

\begin{abstract}
The application of activity recognition in the ``AI + Education" field is gaining increasing attention. However, current work mainly focuses on the recognition of activities in manually captured videos and a limited number of activity types, with little attention given to recognizing activities in surveillance images from real classrooms. In real classroom settings, normal teaching activities such as reading, account for a large proportion of samples, while rare non-teaching activities such as eating, continue to appear. This requires a model that can learn non-teaching activities from few samples without forgetting the normal teaching activities, which necessitates few-shot continual learning (FSCL) capability. To address this gap, we constructed a continual learning dataset focused on classroom surveillance image activity recognition called ARIC (Activity Recognition in Classroom). The dataset has advantages such as multiple perspectives, 32 activity categories, and real-world scenarios, but it also presents challenges like similar activities and imbalanced sample distribution. To overcome these challenges, we designed a few-shot continual learning method that combines supervised contrastive learning (SCL) and an adaptive covariance classifier (ACC). The SCL improves the generalization ability of the model, while the ACC module provides a more accurate description of the distribution of new classes. Experimental results show that our method outperforms other existing approaches on the ARIC dataset.
\end{abstract}


\section{Introduction}
\label{Introduction}
In recent years, activity recognition has gained increasing attention as a significant application of AI in classroom settings. However, existing studies\cite{jisi2021newclass,sharma2024starclass} have primarily focused on the recognition of a limited number of activities, and the data collected are often manually captured videos rather than classroom surveillance images. Activity recognition in classroom surveillance images faces multiple challenges, including class imbalance, high activity similarity, and privacy protection. To fill this gap, we constructed the ARIC dataset, specifically designed for activity recognition in classroom surveillance images. This dataset offers a rich variety of activity types, provides multi-perspective surveillance images, and is sourced from real classroom surveillance videos. However, the ARIC dataset also presents several challenges: 1) an imbalanced distribution of activity categories with significant differences in sample sizes; 2) high similarity between samples of different categories, which can lead to confusion; 3) features extracted by a shallow network to protect privacy, increasing recognition difficulty; and 4) the continuous occurrence of non-instructional activity in real scenarios, requiring the model to have continual learning capabilities.

To address the challenges faced by the ARIC dataset, we can apply few-shot continual learning methods. Few-shot continual learning has garnered significant attention in recent years, with mainstream approaches involving training a feature extractor during the base phase and freezing it during the incremental phase, using class prototypes as classifiers. The FACT\cite{zhou2022forwardfact} method creates virtual classes to reserve space for future classes, SAVC\cite{song2023learningsavc} introduces contrastive learning during base phase and achieves better model generalization through the fantasy space, and ALICE\cite{peng2022fewalice} uses angular penalty loss to achieve more compact intra-class clustering.

Nevertheless, current methods remain inadequate in addressing the specific challenges posed by the ARIC dataset. To this end, we propose a specialized few-shot continual learning method for activity recognition in classroom surveillance images. During the base phase, we use a feature-augmented supervised contrastive learning approach to enhance the model's generalization ability and reserve space for future activity categories to better achieve future class predictions. In the incremental phase, the covariance matrix is used as a memory unit, combined with an adaptive mechanism to form the ACC module. By analyzing the variance of new classes, it dynamically adjusts the classifier's decision boundaries to match the feature distribution of the new classes, effectively addressing the issues of small sample size and similarity between new and old classes. Experimental results demonstrate that our method outperforms existing approaches on the ARIC dataset.

\section{ARIC-Dataset}
\label{ARIC-Dataset}
\begin{figure}[htbp]
    \centering
    \subfigure[front]{\includegraphics[width=0.4\hsize, height=0.25\hsize]{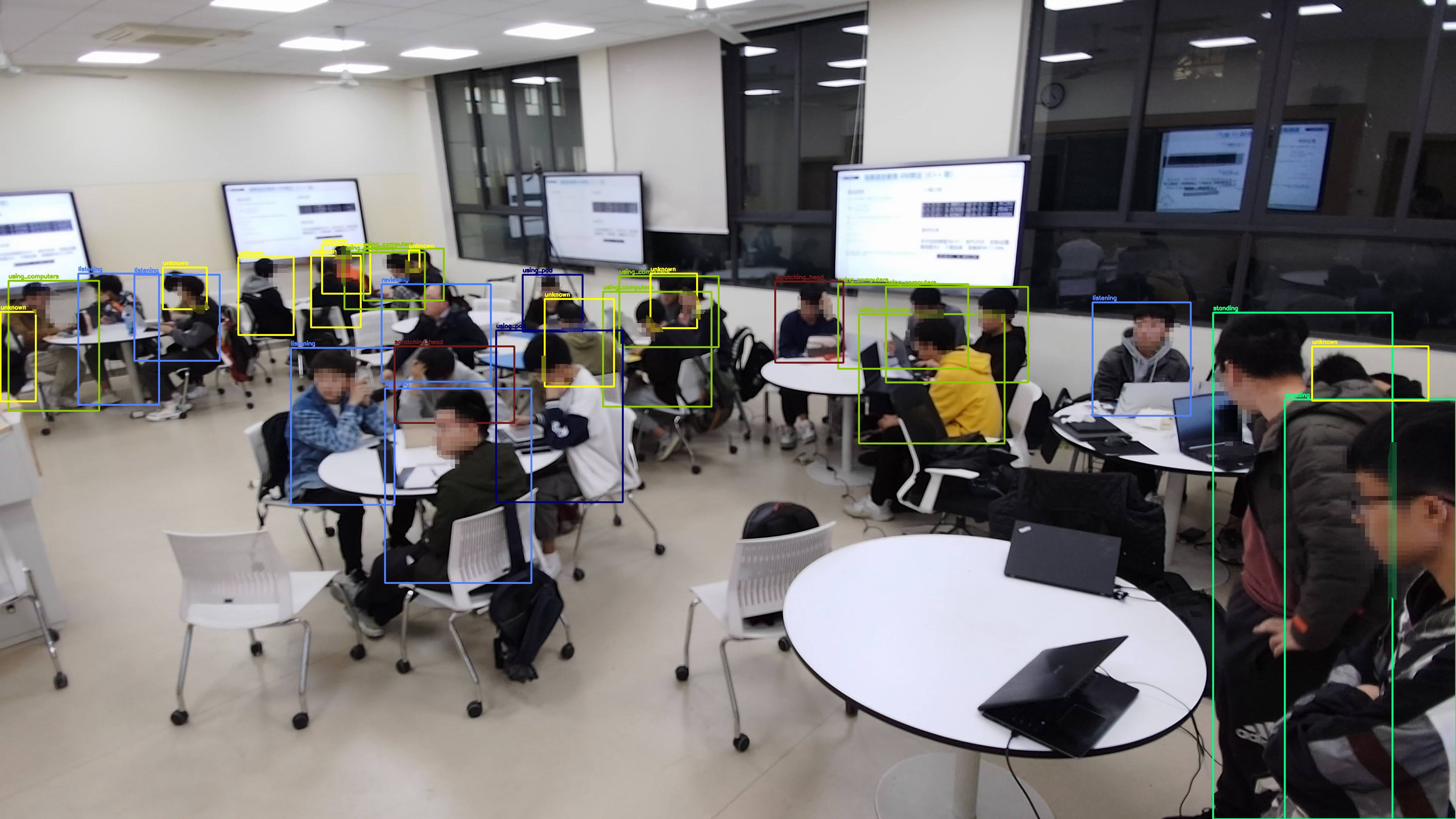}\label{data_1}}
    \subfigure[middle]{\includegraphics[width=0.4\hsize, height=0.25\hsize]{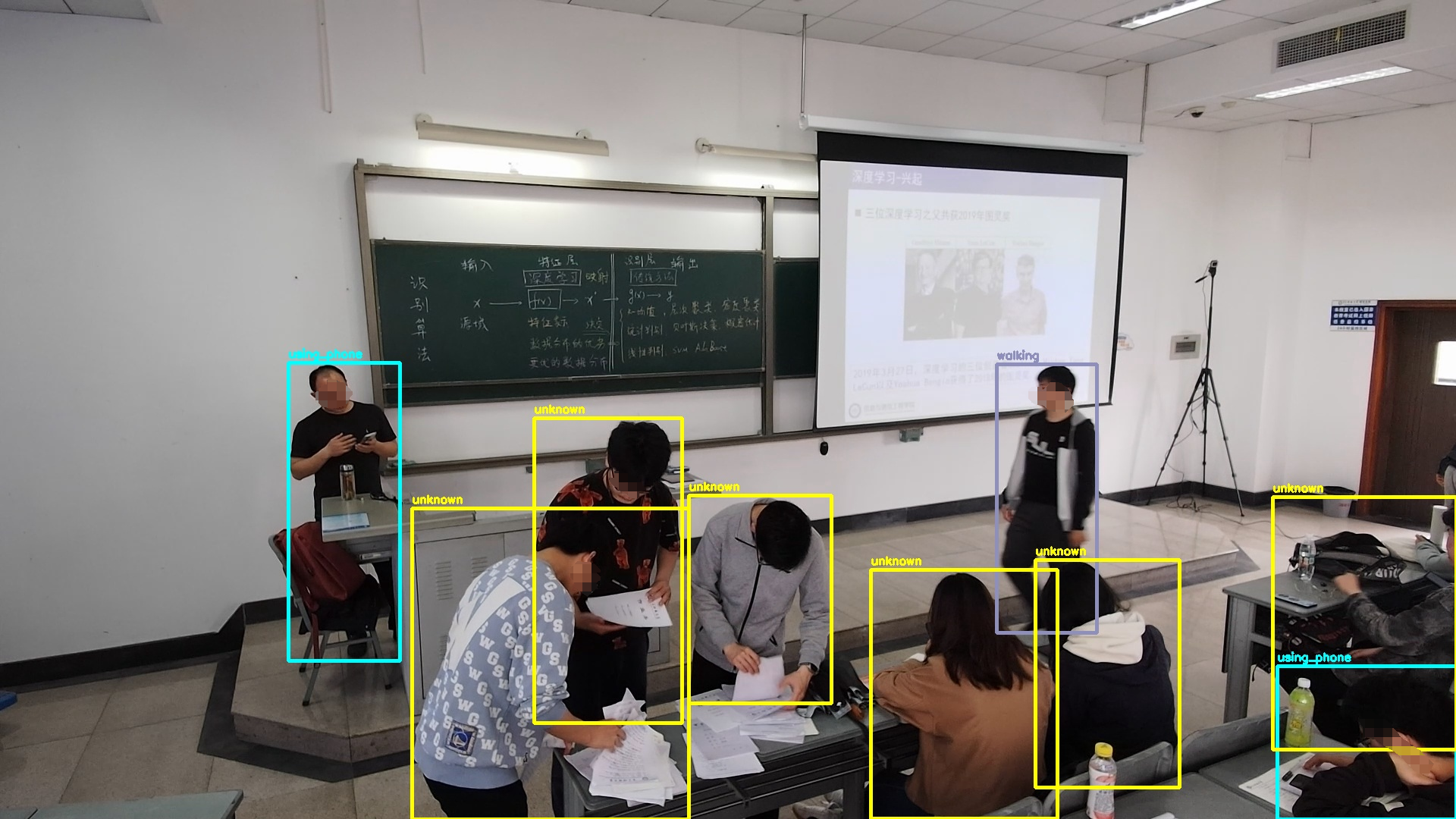}\label{data_2}}
    \subfigure[rear]{\includegraphics[width=0.4\hsize, height=0.25\hsize]{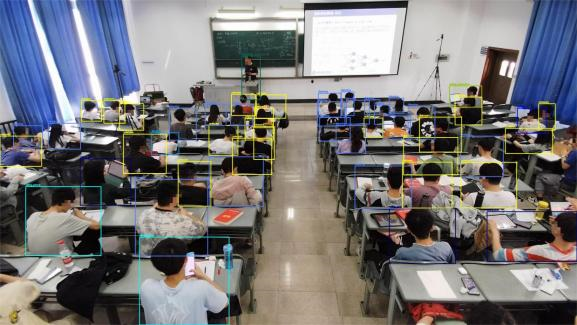}\label{data_3}}
    \caption{\small{Monitoring samples from different perspectives.}
    \label{dataset}}
\end{figure}

\begin{figure}[htbp]
    \centering
    \includegraphics[width=\linewidth]{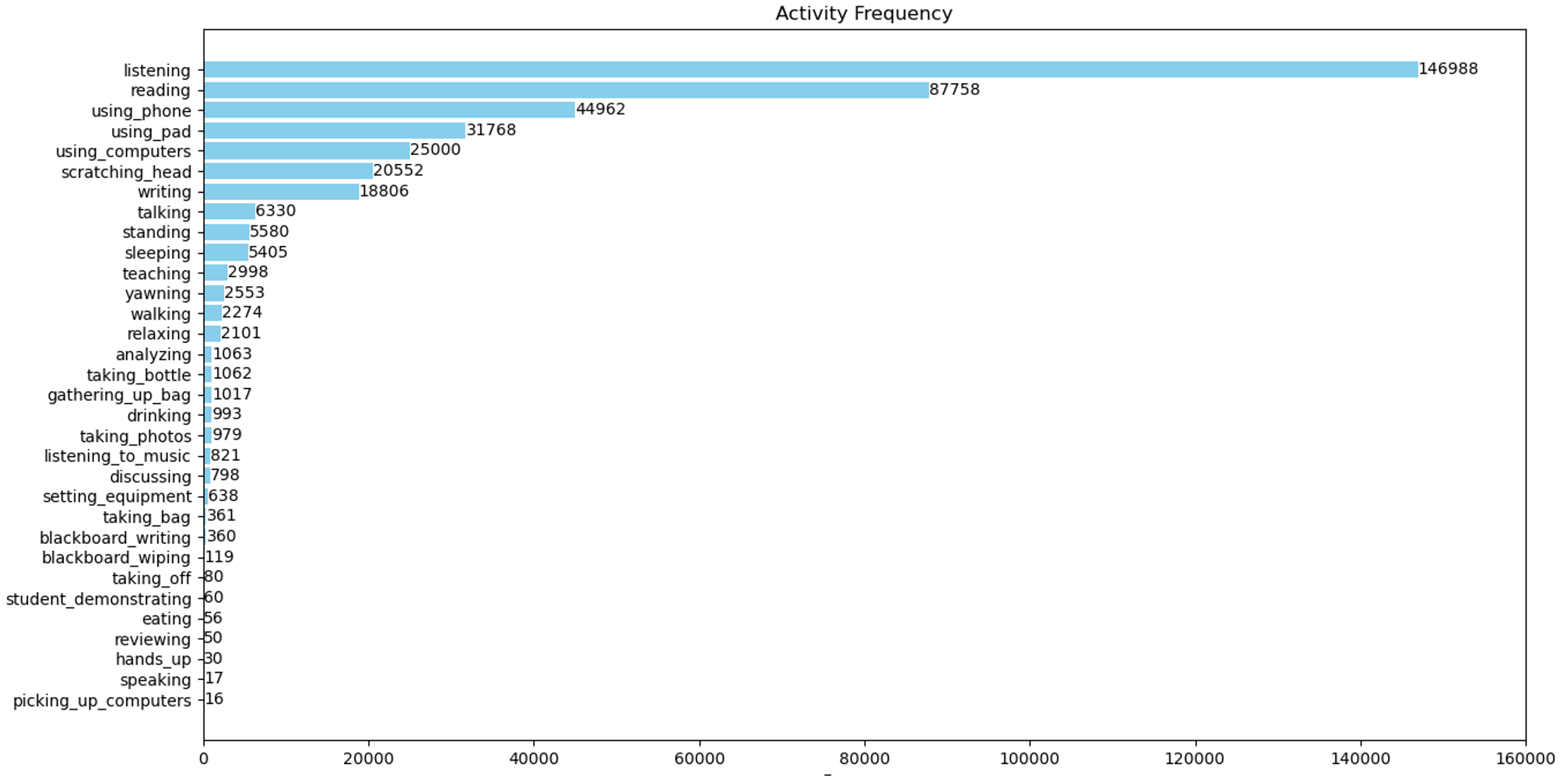}
    \vspace{-1em}
    \caption{\small{Sample distribution of the 32 activity categories.}}
    \label{dis}
\end{figure}

The ARIC is a brand-new and challenging dataset based on real classroom surveillance scenarios. We used surveillance videos from three different perspectives—front, middle, and real—of real classroom scenarios as the raw data.(as shown in Fig. \ref{dataset}). Images were then extracted from these videos, and the activities of students and teachers within the images were annotated, forming the image modality. We also extracted audio corresponding to 5 seconds before and after each image (a total of 10 seconds) as the audio modality. Additionally, we used the open-source large model InternVL\cite{chen2024internvl} to generate captions for each image as the text modality.
The ARIC dataset is characterized by its real classroom scenarios, three modalities, and diverse perspectives. The complexity of human activities, the diversity of actions, and the uniqueness of crowded classroom scenes make this dataset highly challenging. The dataset consists of 36,453 surveillance images covering 32 classroom activities, such as listening to lecture, reading, and using mobile phone. The distribution of samples across different activities is shown in Fig. \ref{dis}.

To protect the privacy of individuals appearing in the images and to avoid releasing the original images, we used shallow layers of pre-trained models to convert the original images into feature data. Considering the need for backbone models in the field of continual learning, we selected three commonly used pre-trained models: ResNet50\cite{he2016deep(resnet)}, ViT\cite{dosovitskiy2020image(vit)}, and CLIP-ViT\cite{radford2021learning(clipvit)}. For example, by using conv1 layer and 3x3 max pool layer of the ResNet50 pre-trained model, we converted the image data into feature data with dimensions of [1, 64, 56, 56].

We also pre-defined reasonable incremental learning task divisions within the dataset to standardize experiments across the dataset: A) In the base phase, provide a few categories with a large number of samples, then randomly and as evenly as possible distribute the remaining categories across different incremental phases. B) Arrange the categories in descending order by the number of samples and then allocate them to different incremental phases based on this order. The specific partitioning schemes will be represented using the formula: $\boldsymbol{B}+\boldsymbol{S}\times \boldsymbol{N}$. Here, $\boldsymbol{B}$ represents the number of the base class, $\boldsymbol{S}$ represents the number of incremental phases, and $\boldsymbol{N}$ represents the number of categories in each incremental phase. For example, $8+6\times4$ means there are $8$  base categories, $6$ incremental phases, and $4$ categories in each incremental phase. 

The ARIC dataset can be downloaded, and more detailed information can be obtained by the link: \href{https://ivipclab.github.io/publication_ARIC/ARIC}{https://ivipclab.github.io/publication\_ARIC/ARIC.}

\begin{figure*}[htp]
    \centering
    \includegraphics[width=14cm]{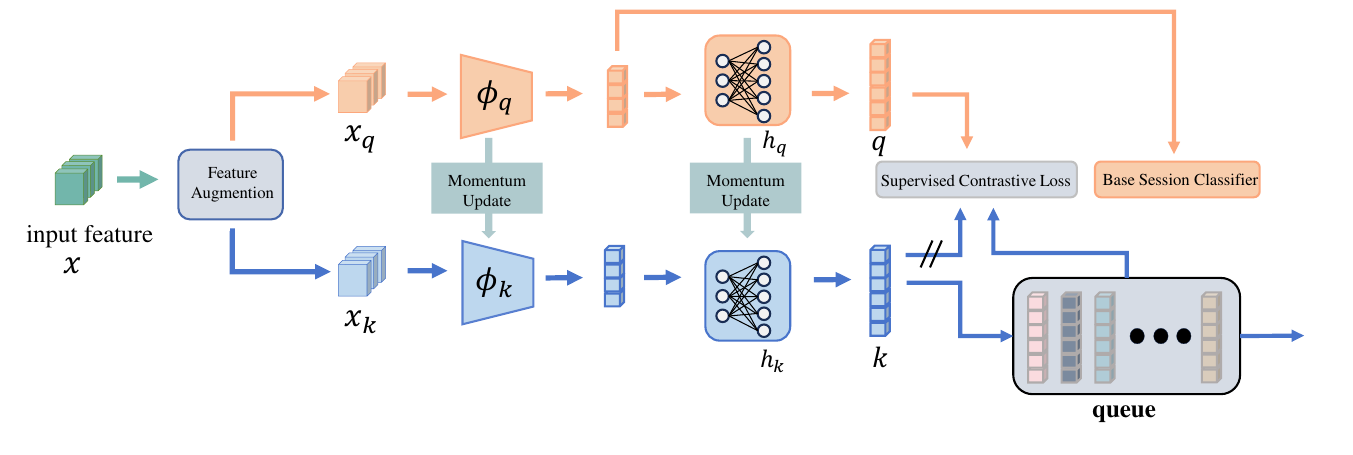}
    \vspace{-1em}
    \caption{\small{Pipline of feature-augmented supervised contrastive learning.}}
    \label{F1}
\end{figure*}

\section{Method}
\label{sec:typestyle}
In this section, we will first introduce the task setup for FSCL, followed by an explanation of our proposed method.

\subsection{Few-Shot Continual Learning}
\label{ssec:Few-Shot Continua Learning}

\textbf{Base Session:} In FSCL, the dataset needs to provide a base class training set with sufficient samples, denoted as \( \mathcal{D}^{0} = \left\{ \left(\mathbf{x}_i, \mathbf{y}_i\right)\right\}_{i=1}^{N_0} \), and a base class test set \( \mathcal{D}_{t}^{0} = \left\{ \left(\mathbf{x}_i, \mathbf{y}_i\right)\right\}_{i=1}^{M_0} \), where \( N_0 \) and \( M_0 \) represent the number of samples in the training set and test set respectively. Here, \( \mathbf{x}_i \in \mathbb{R}^D \) is the training instance for \( \mathbf{y}_i \in Y_0 \), and \( Y_0 \) is the label space of the base task.

\textbf{Incremental Session:} In this stage, the training set for new tasks \( \left\{ \mathcal{D}^{1} ,\dots,\mathcal{D}^{B}\right\} \) are introduced sequentially. Each set is denoted as \( \mathcal{D}^{b} = \left\{ \left(\mathbf{x}_i, \mathbf{y}_i\right)\right\}_{i=1}^{N_b} \), where \( \mathbf{y}_i \in Y_b \), and \( Y_b \cap Y_{b'} = \varnothing \text{ for } b \neq b' \). The dataset \( \mathcal{D}^b \) is only accessible during the training phase of task \( b \). The limited instances in each dataset can be organized in an \( N \)-way, \( K \)-shot format, representing \( N \) classes with \( K \) sample instances per class at each incremental stage.


\subsection{Feature-Augmented Supervised Contrastive Learning}
\label{ssec:Feature-Augmented Supervised Contrastive Learning}

To address the challenge of high similarity between different activities in the ARIC dataset, we introduce supervised contrastive learning during the base phase. SCL is particularly effective in handling fine-grained differences, enabling the model to better distinguish and amplify subtle variations\cite{10.5555/3524938.3525087} between easily confused categories, such as reading a book versus looking at a phone. Additionally, SCL contributes to achieving more compact clustering, which reserves space for future incremental categories and thus enhances the model’s ability for FSCL.

In contrastive learning, image augmentation techniques play a crucial role\cite{zbontar2021barlow,grill2020bootstrap}. However, since the ARIC dataset is released as features rather than images, we designed a feature augmentation strategy that adapts traditional image augmentation methods, including cropping, flipping, and rotation, to the feature space. This strategy is integrated into the MoCo\cite{he2020momentum} framework to implement SCL, as shown in Fig. \ref{F1}. This framework maintains a continuously updated feature repository, allowing the model to learn the most recent feature representations. In each training iteration, we first apply a series of random augmentations to the input feature $\mathbf{x}$, generating two augmented views $\mathbf{x}_q$ and $\mathbf{x}_k$. These are then processed by their respective encoders $\phi_q$, $\phi_k$ and projection layers $h_q$, $h_k$, resulting in query feature $\mathbf{q}$ and key feature $\mathbf{k}$. A feature queue stores the most recently computed key features along with their label information. The key network is updated using a momentum mechanism to ensure smoother and more robust parameter updates. This setup enables the model to learn more discriminative feature representations from a large pool of samples.

The supervised contrastive loss for each feature sample $\mathbf{x}$ is computed as follows:

\begin{equation}
  \begin{aligned}
&\mathcal{L}_{\text {SCL }}(\mathbf{x}) \\ 
&= -\frac{1}{|P(\mathbf{x})|} \sum_{\mathbf{k}_{+} \in P(\mathbf{x})} \log \frac{\exp \left(\mathbf{q}\cdot \mathbf{k}_{+} / \tau\right)}{\sum_{\mathbf{k}^{\prime} \in \mathbf{k} \cup \mathbf{Q} } \exp \left(\mathbf{q}\cdot \mathbf{k}^{\prime} / \tau\right)}
\end{aligned}  
\end{equation}

Here, $\mathbf{Q}$ represents the feature queue, and $P(\mathbf{x})$ denotes the set of positive samples, which is the set of samples in $\mathbf{k} \cup \mathbf{Q}$ that belong to the same class as $\mathbf{x}$.

During the base phase, in addition to the SCL loss, we also use a cross-entropy classification loss to simultaneously optimize the model's classification ability and the discriminability of feature representations. We use $\phi_q$ in the query network as a feature extractor to extract features for classification. The cross-entropy classification loss is defined as follows:

\begin{equation}
  \mathcal{L}_{\text{cls}}(\mathbf{x}, \mathbf{y}) = \mathcal{L}_{\text{ce}}(W^{\top} \phi_q(\mathbf{x}), \mathbf{y})  
\end{equation}

where $\mathcal{L}_{\text{ce}}(\cdot, \cdot)$ denotes the cross-entropy loss, $W \in \mathbb{R}^{d \times |Y_0|}$, and $\phi_q(\mathbf{x}) \in \mathbb{R}^{d \times 1}$.

The final loss function is:
\begin{equation}
    \mathcal{L}_{\text{total}} = \mathcal{L}_{\text{cls}} +  \mathcal{L}_{\text{SCL}}
\end{equation}

\begin{figure}[t]
    \centering
    \includegraphics[width=\linewidth]{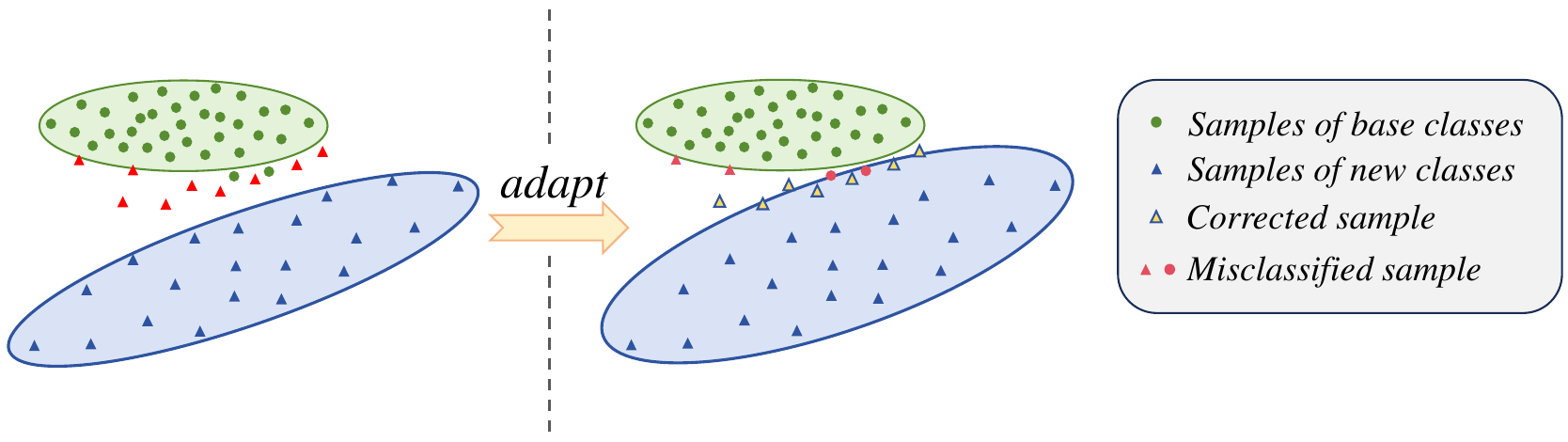}
    \vspace{-1em}
    \caption{\small{Qualitative illustration of the adaptive mechanism. Circles represent samples from old classes, and triangles represent samples from new classes. Red circles and triangles indicate misclassified samples, while yellow triangles represent corrected new class samples.}}
    \label{F2}
\end{figure}

\subsection{Adaptive Covariance Classifier }
\label{Adaptive Covariance Classifier (ACC)}
Traditional classifiers based on the Nearest Class Mean (NCM) rely on learning features from all classes together. However, in incremental learning, dynamic data streams can make NCM less effective. Mensink et al.\cite{mensink2013distance} introduced the use of Mahalanobis distance to measure the distance between samples and classes, which is better suited for this scenario\cite{goswami2024fecam}. Additionally, a feature extractor trained only on base classes can result in high semantic similarity between new classes and some old classes\cite{wang2024few}. As shown on the left side of Fig. \ref{F2}, some new class samples have features that are too close to old classes, leading to classification errors. Our proposed ACC module leverages class variance characteristics to adjust the covariance matrix, making it more aligned with the class feature distribution. After adjustment, the decision boundaries for the new classes, as shown on the right side of the Fig. \ref{F2}, allow a significant portion of the new classes to be correctly reclassified.

When predicting the label of a sample, the Mahalanobis distance $ \mathbf{D}(\mathbf{x}) $ is used to calculate the distance between the sample and the class. Here, $ \mathbf{G} $ represents the Gaussian-transformed feature vector of the sample $ \mathbf{x} $, denoted as $\mathbf{G}(\phi_q(\mathbf{x}))$, and $\boldsymbol{\mu} $ is the mean vector of the class, while $\mathbf{\Sigma_{a}}$ is the adaptive covariance matrix.

\begin{equation}
\mathbf{D}(\mathbf{x}) = \sqrt{(\mathbf{G} -\boldsymbol{\mu})^\top \mathbf{\Sigma_{a}}^{-1} 
 (\mathbf{G} - \boldsymbol{\mu})}
\end{equation}

Using Gaussian-transformed data helps generate representative samples, but raw feature data often exhibits skewness\cite{yang2021free}. To ensure that the input features approximate a Gaussian distribution, we applied the Box-Cox transformation, where $ \lambda $ is a hyperparameter:

\begin{equation}
    \mathbf G(x) = 
    \begin{cases} 
    \frac{x^\lambda - 1}{\lambda} & \text{if } \lambda \neq 0 \\
    \log(x) & \text{if } \lambda = 0 
    \end{cases}
\end{equation}

In few-shot learning scenarios, the number of samples is much smaller than the feature dimensions, which can result in a rank-deficient covariance matrix, making it impossible to compute its inverse. To address this, we introduced covariance shrinkage\cite{kumar2022gdc}, incorporating the adaptive parameter $ \alpha $, and applied normalization to compute the adaptive covariance matrix $ \mathbf{\Sigma_{a}} $. 

\begin{equation}
    \mathbf{\Sigma_{a}} = Normal[ \mathbf{\Sigma} + \alpha \sigma_1 \mathbf{I} +  \sigma_2 (\mathbf{1} - \mathbf{I})]
\end{equation}

\begin{equation}
 \alpha = \frac{k}{{N_b}} \sum_{i=1}^{{N_b}} \left(\phi_q(\mathbf{x}) - \boldsymbol{\mu}\right)^{2} 
\end{equation}

Here, $ \mathbf{\Sigma} $ is the class covariance matrix, $ \mathbf{I} $ is an identity matrix of the same shape as $ \mathbf{\Sigma} $, and $ \mathbf{1} $ is an all-ones matrix of the same shape as $ \mathbf{\Sigma} $. The values $ \sigma_1 $ and $ \sigma_2 $ represent the mean of the diagonal and off-diagonal elements of $ \mathbf{\Sigma} $, respectively, with a scaling factor $ k > 1 $. The adaptive parameter $ \alpha $ adjusts the covariance matrix through $ \sigma_1 $ and $ \sigma_2 $.

\section{Experiments}
\label{sec:Experiments}

\subsection{Implementation Details}
\label{ssec:Implementation Details}
We evaluate our proposed method on the ARIC dataset, using only the image modality for this experiment. The task is divided as follows: the base phase utilizes 20 classes, and in the subsequent 4 incremental phases, 3 new classes are introduced at each phase, following a 3-way 5-shot setting. In each incremental phase, only 5 samples per new class are provided for training. This setup simulates the scenario in classroom surveillance images where non-instructional activities continuously appear but with a limited number of samples, requiring the model to learn effectively under constrained sample conditions. We adopt ResNet18\cite{he2016deep(resnet)} as the backbone of our network. In base phase, we utilized an SGD optimizer with a momentum of 0.9 and an initial learning rate of 0.1, adjusted using a cosine annealing scheduler. The model is trained for 200 epochs with a batch size of 256. During the incremental phase, set the parameter $\lambda=0.2$ for the Gaussian transformation, and set the adaptive scaling factor $k=4$ in the ACC module.

\begin{table}[t]
\caption{\small{Average TOP-1 accuracy at different stages of incremental tasks on the ARIC dataset. (Best results are highlighted in bold.)}}
\vspace{0.5em}
\centering
\small
\renewcommand\arraystretch{1}
\resizebox{.95\linewidth}{!}{
    \begin{tabular}{l|ccccc}
    \toprule[1pt]

     \multirow{2}{*}{Method}    &\multirow{2}{*}{0}  &\multirow{2}{*}{1} &\multirow{2}{*}{2} &\multirow{2}{*}{3} &\multirow{2}{*}{4} \\
    \vspace{0.1em} \\
    \hline
    Finetune & 51.7 & 7.67 & 5.55 & 2.21 & 1.23 \\ 
    Teen\cite{wang2024few} & 52.6 & 47.33 & 44.87 & 40.77 & 40.12 \\ 
    ALICE\cite{peng2022fewalice} & 61.1 & 52.72 & 49.89 & 47.21 & 44.07 \\ 
    FACT\cite{zhou2022forwardfact} & 66.7 & 59.60 & 55.71 & 53.19 & 46.33 \\ 
    SAVC\cite{song2023learningsavc} & \textbf{68.13} & 63.35 & 60.18 & 56.79 & 53.41 \\ 
    Our & 67.6 & \textbf{64.54} & \textbf{61.7} & \textbf{57.97} & \textbf{55.95} \\ 
    
    \bottomrule[1pt]
    \end{tabular}
}
\label{imfe}
\end{table}

\subsection{Result}
\label{ssec:Result}
The experimental results of our method on the ARIC dataset are shown in Table \ref{imfe} (the evaluation metric is the average TOP-1 accuracy tested on all known classes). To demonstrate the effectiveness of our method on the ARIC dataset, we compared it with several state-of-the-art few-shot continual learning methods. ALICE\cite{peng2022fewalice}, FACT\cite{zhou2022forwardfact}, and SAVC\cite{song2023learningsavc}, are all based on the feature space and use prototypes as classifiers, while Teen\cite{wang2024few} only adjusts the prototype classifier during the incremental phase. Additionally, we present the results of Finetune without using any continual learning methods. The experimental results show that our proposed method significantly outperforms existing methods in each incremental task on the ARIC dataset.

\subsection{Ablation Study}
\label{ssec:Analysis}
To evaluate the impact of each component in our proposed method, we conducted ablation experiments, as shown in Table \ref{ablation}. First, when we disabled the SCL loss ($ \mathcal{L}_{\text{SCL}} $) during the base phase, the experimental results showed a significant drop in the model's performance on base class classification. This indicates that SCL allows the model to more accurately distinguish between easily confused categories. Additionally, the model’s performance after the last incremental phase also declined, suggesting that SCL achieved more compact clustering during the base phase, thereby enhancing the model’s few-shot continual learning ability. Second, when we removed the ACC module and used only the prototype classifier, the experimental results showed a decrease in performance at each incremental stage, indicating that the ACC classifier better defines the decision boundaries for each class in incremental tasks. We also added the ACC module to the FACT method for experimentation, and the results similarly demonstrated this point.

Finally, to verify the impact of the adaptive mechanism on the ACC module, we fixed the adaptive parameter $ \alpha $ to 1 in our experiments. The results, shown in Table \ref{ablation_2}, indicate that disabling the adaptive mechanism led to a decline in performance across all incremental stages. This demonstrates the significant improvement provided by the adaptive mechanism to the covariance classifier, offering a quantitative assessment of its contribution to the model's performance.

\begin{table}[ht]
\caption{\small{Ablation study results for different components of our method. (Best results are highlighted in bold.)}}
\vspace{0.5em}
\centering
\small
\renewcommand\arraystretch{1}
\resizebox{.95\linewidth}{!}{
    \begin{tabular}{l|ccccc}
    \toprule[1pt]
    \multirow{2}{*}{Method}    &\multirow{2}{*}{0}  &\multirow{2}{*}{1} &\multirow{2}{*}{2} &\multirow{2}{*}{3} &\multirow{2}{*}{4} \\
    \vspace{0.1em} \\
    \hline
    FACT     &66.7 &59.6 &55.714 &53.19 &46.325 \\
    FACT $w/ \, ACC$ &66.6 &61.72 &57.6 &53.5 &48.8 \\
    Our $w/o \, ACC$ &67.6 &64.28 &60.21 &56.5 &53.89 \\
    Our $w/o \, \mathcal{L}_{\text{SCL}}$ &64.25 &60.59 &56.71 &53.34 &50.11 \\   
    Our &\textbf{67.6} &\textbf{64.54} &\textbf{61.7} &\textbf{57.97} &\textbf{55.95}\\
    
    \bottomrule[1pt]
    \end{tabular}
}
\label{ablation}
\end{table}

\begin{table}[ht]
\caption{\small{Ablation study results of the adaptive mechanism in the ACC module. }}
\vspace{0.5em}
\centering
\small
\renewcommand\arraystretch{1}
\resizebox{.95\linewidth}{!}{
    \begin{tabular}{l|ccccc}
    \toprule[1pt]
    \multirow{2}{*}{Method}    &\multirow{2}{*}{0}  &\multirow{2}{*}{1} &\multirow{2}{*}{2} &\multirow{2}{*}{3} &\multirow{2}{*}{4} \\
    \vspace{0.1em} \\
    \hline
    Our $\alpha=1$ &67.6 &64.42 &60.7 &56.81 &54.39 \\   
    Our &\textbf{67.6} &\textbf{64.54} &\textbf{61.7} &\textbf{57.97} &\textbf{55.95}\\
    
    \bottomrule[1pt]
    \end{tabular}
}
\label{ablation_2}
\end{table}

\subsection{Visualization}
\label{sec:Visualization}
\begin{figure}[!htb]
    \centering
    \subfigure[Fintune]{\includegraphics[width=0.3\hsize, height=0.3\hsize]{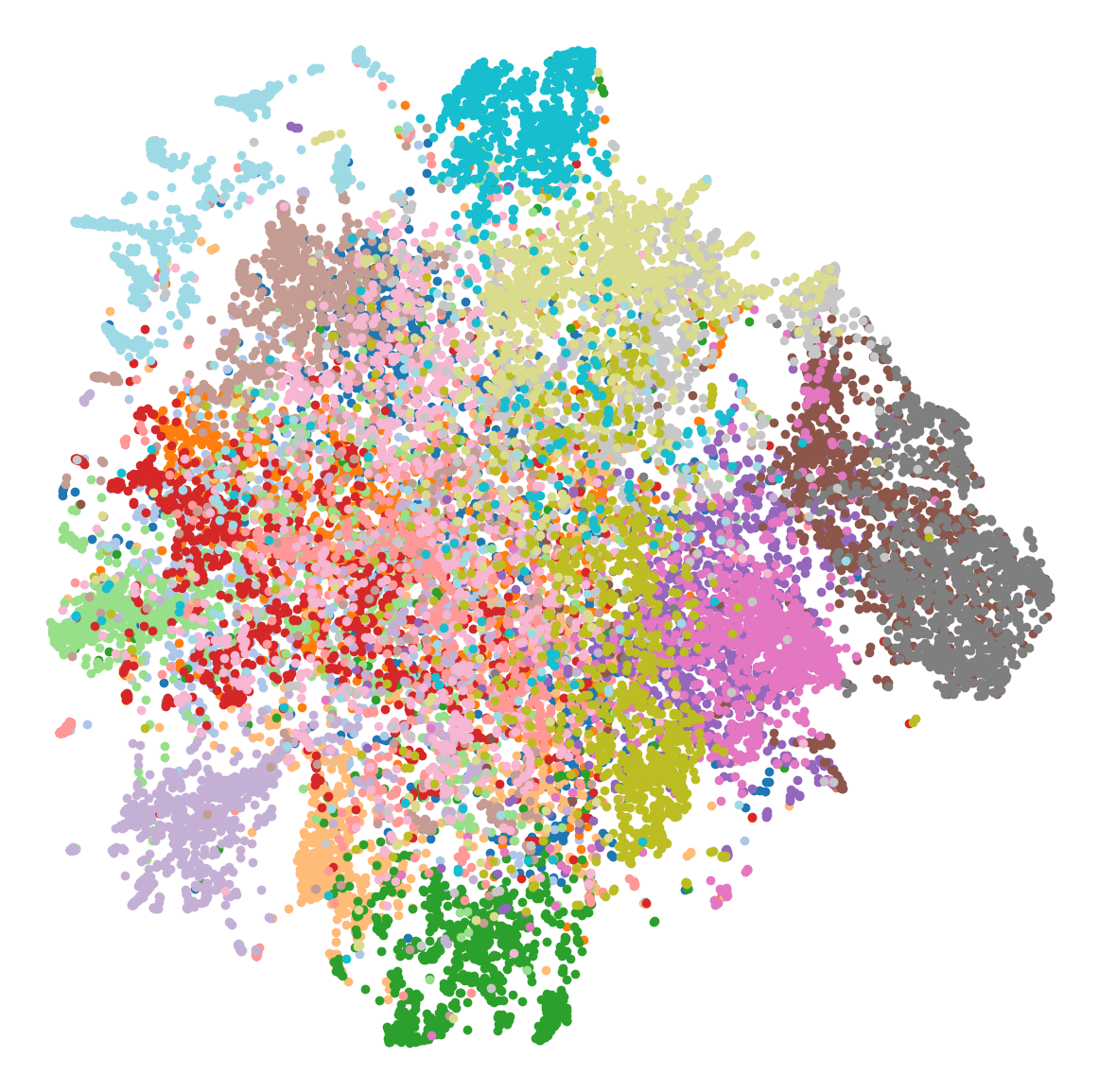}\label{fig: sub_figure1}}
    \subfigure[FACT]{\includegraphics[width=0.3\hsize, height=0.3\hsize]{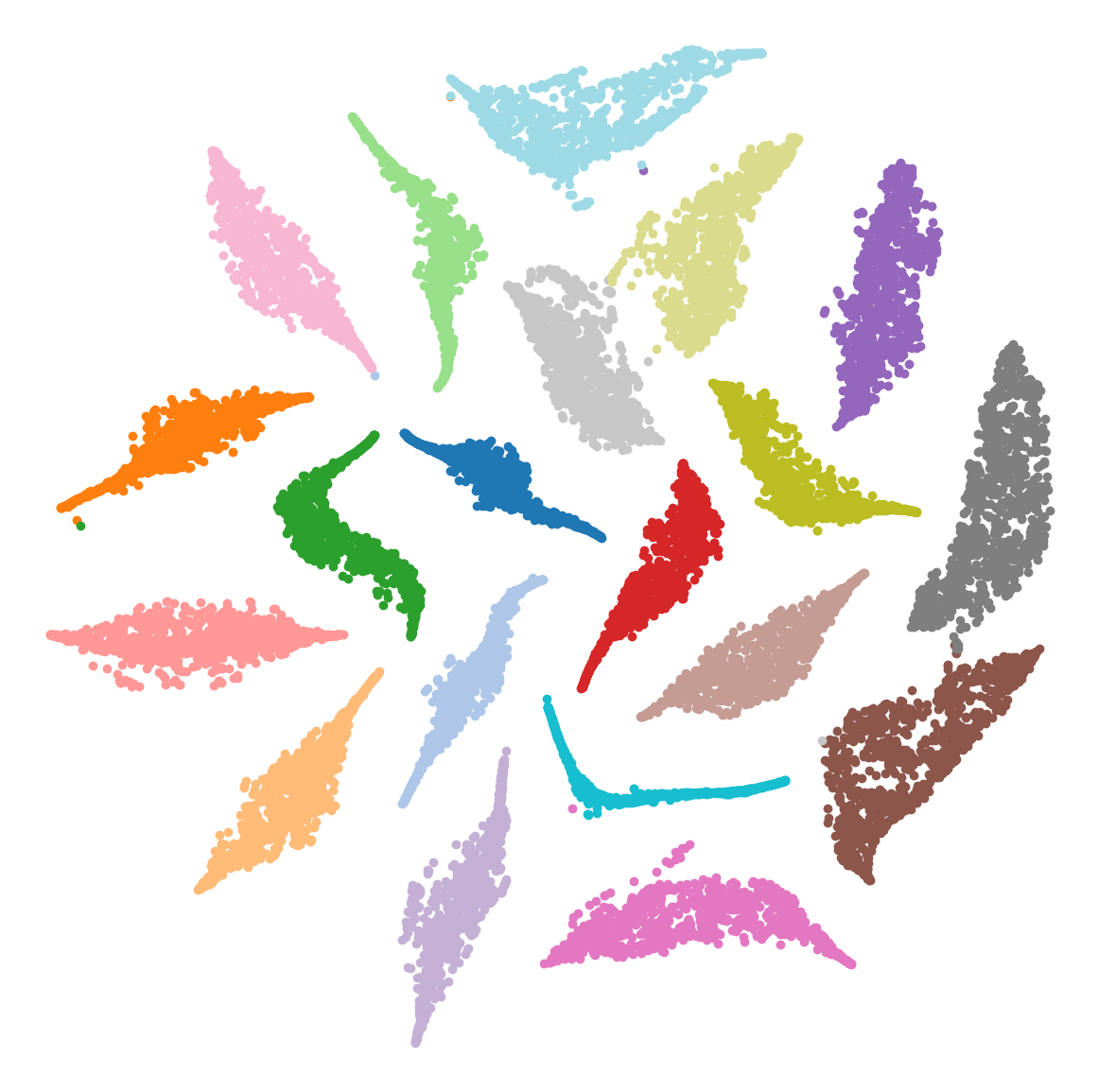}\label{fig: sub_figure2}}
    \subfigure[Our]{\includegraphics[width=0.3\hsize, height=0.3\hsize]{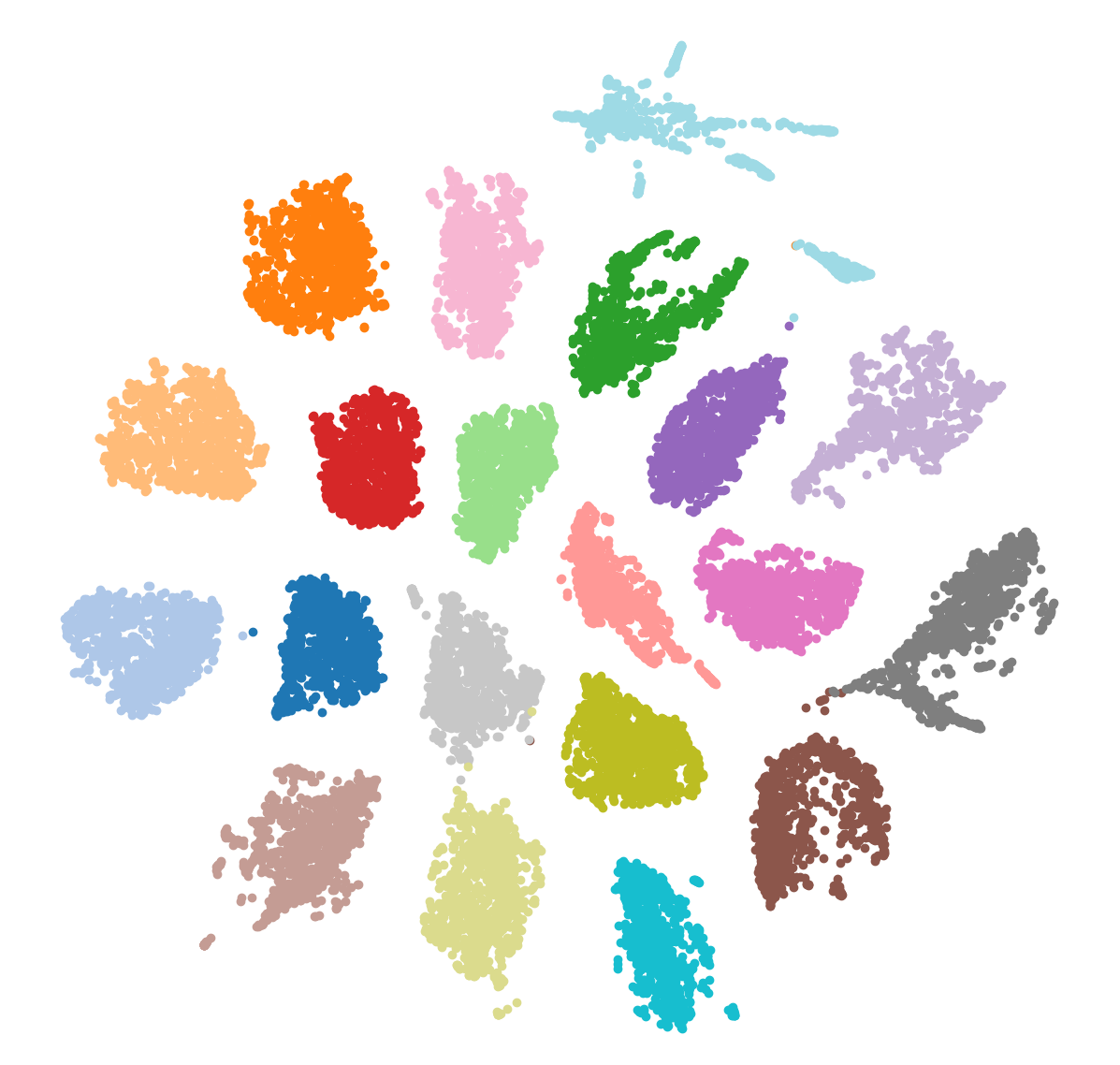}\label{fig: sub_figure3}}
    \caption{The t-SNE plots of base class feature distributions after base stage training on the ARIC dataset for different methods: (a) Finetune, (b) FACT, (c) Our.}
    \label{F3}
\end{figure}

As illustrated in Fig. \ref{F3}, we compared the feature distribution in the feature space after the base phase for three different methods. It is evident that our method achieves superior clustering for the base classes. Compared to Fintune and FACT, incorporating supervised contrastive learning effectively increases the inter-class distance while reducing the intra-class distance, resulting in more compact clusters for each class. This significantly contributes to accurate base class recognition and better integration of incremental classes into the feature space.

\section{Conclusion}
\label{sec:Conclusion}
In this study, we tackled the unique challenges of activity recognition in classroom surveillance images, particularly those presented by the ARIC dataset, by developing an innovative few-shot continual learning method. Our approach effectively addresses issues such as class imbalance, high activity similarity, and the need for privacy-preserving features. This is achieved by integrating feature-augmented SCL in the base phase and the ACC module in the incremental phase. The experimental results on the ARIC dataset demonstrate that our method significantly enhances the model's generalization ability and improves classifier accuracy.

\newpage
\bibliographystyle{IEEEtran}
\bibliography{refs}

\end{document}